\documentclass[10pt,twocolumn,letterpaper]{article}

\usepackage{xcolor}
\usepackage{colortbl}
\usepackage{booktabs}
\usepackage{array}
\usepackage{amsmath}
\usepackage{graphicx}
\usepackage{caption}
\usepackage{placeins}
\usepackage[pagenumbers]{cvpr}
\usepackage[numbers]{natbib}

\usepackage[accsupp]{axessibility}
\definecolor{cvprblue}{rgb}{0.21,0.49,0.74}
\usepackage[pagebackref,breaklinks,colorlinks,allcolors=cvprblue]{hyperref}
\usepackage[capitalize]{cleveref}

\newcommand{\custompar}[1]{
  \par
  \vspace{4pt}
  \noindent\textbf{#1}
}

\title{
Are Pretrained Image Matchers Good Enough\\for SAR--Optical
Satellite Registration?
}

\author{
Isaac Corley$^{1*}$ \quad
Alex Stoken$^{2}$ \quad
Gabriele Berton$^{2}$
\\\\
$^{1}$Taylor Geospatial \quad
$^{2}$Independent Researcher \quad
}

\def\paperID{16}
\def\confName{CVPR}
\def\confYear{2026}

\begin{document}
\setlength{\tabcolsep}{2.5pt}
\setlength{\textfloatsep}{6pt plus 1pt minus 2pt}
\setlength{\floatsep}{6pt plus 1pt minus 2pt}
\setlength{\intextsep}{6pt plus 1pt minus 2pt}
\setlength{\dbltextfloatsep}{6pt plus 1pt minus 2pt}
\setlength{\dblfloatsep}{6pt plus 1pt minus 2pt}
\twocolumn[{%
    \maketitle
    \vspace{-2.0em}
    \begin{center}
      \includegraphics[width=0.99\textwidth]{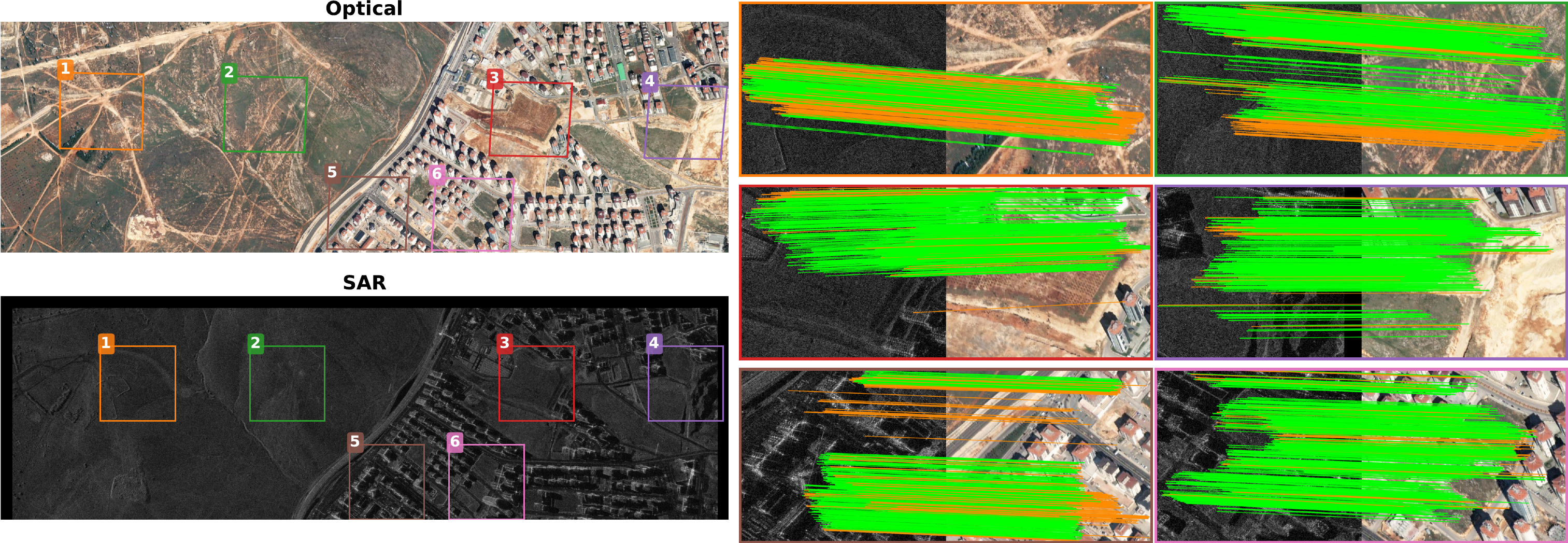}
      \captionsetup{hypcap=false}
      \captionof{figure}{\textbf{Zero-shot SAR--optical registration
        on SpaceNet9.}
        Left column: full-resolution optical scene (top) and full-resolution SAR scene (bottom).
        Right column: tiled correspondence gallery from
        a zero-shot pretrained matcher (XoFTR)---each panel shows one randomly sampled,
        tie-point-projected overlapping region from the same geographic area in both modalities.
        Green lines denote affine-RANSAC inliers; orange lines denote outliers.
        Without satellite-domain fine-tuning or adaptation, XoFTR
      produces dense inlier fields on SAR--optical pairs.}
      \label{fig:hero}
    \end{center}
    \vspace{0.4em}
}]

\begin{abstract}
Cross-modal optical--SAR (Synthetic Aperture Radar) registration is a bottleneck in remote-sensing disaster response. Modern image matchers are developed and benchmarked almost exclusively on natural-image domains. We evaluate twenty-four pretrained matcher configurations in a zero-shot setting, with no fine-tuning or domain adaptation on satellite or SAR data. The evaluation spans SpaceNet9 and two additional cross-modal benchmarks under a deterministic protocol that uses tiled large-image inference, robust geometric filtering, and tie-point-grounded metrics. Our results show uneven transfer: matchers with explicit cross-modal training do not uniformly outperform those without it. XoFTR (trained for visible--thermal matching) and RoMa achieve the lowest reported mean tie-point error at $3.0$\,px on the labeled SpaceNet9 training scenes. RoMa achieves this result \emph{without any cross-modal training}.
MatchAnything-ELoFTR ($3.4$\,px), trained on synthetic cross-modal pairs, is close behind. XoFTR also runs roughly an order of magnitude faster than the RoMa family (${\approx}0.4$\,s vs.\ $5.0$\,s per pair). RoMa's performance is consistent with the hypothesis that frozen DINOv2 features confer robustness to large appearance shifts. Deployment protocol choices (geometry model, tile size, inlier gating) change mean error by up to $33\times$ for a single matcher. This shift can exceed the effect of swapping matchers within our evaluated protocol ablations. Affine geometry alone reduces mean error from $12.3$ to $9.7$\,px.
\end{abstract}

\section{Introduction}
 \footnotetext[1]{Corresponding author: isaac.corley@taylorgeospatial.org}

When a hurricane makes landfall, cloud cover often renders optical imagery
unusable. SAR then becomes the primary observation modality.
Registering newly acquired SAR images to pre-existing optical basemaps is
necessary to produce georeferenced damage assessments within hours.
Pretrained image matchers are natural candidates for this task.
Yet the best pretrained image matchers are designed for indoor and urban
natural-image benchmarks, not for overhead cross-modal imagery with speckle,
layover, and radiometric inversion.
As Robinson~\etal~\citep{robinson2026advancing} observe, ``using satellite imagery is not simply
`computer vision with larger images'\,''; Rolf~\etal~\citep{rolf2024mission} make the
stronger claim that satellite data constitutes a \emph{distinct modality} whose statistical
properties---overhead perspective, sensor diversity, global geographic distribution---systematically
violate assumptions baked into natural-image pretraining.
Do pretrained matchers survive this domain shift to overhead optical--SAR
imagery? Or do we need remote-sensing-specific registration methods?

\paragraph{Why is cross-modal satellite registration hard?}
Optical and SAR sensors observe the same scene through fundamentally different
physics.
Optical sensors record reflected sunlight and produce texture-rich imagery with
familiar visual structure; SAR sensors emit and receive microwave pulses,
yielding imagery dominated by speckle noise, geometric layover from
side-looking acquisition, and radiometric inversion relative to optical
data~\citep{hansch2024spacenet9}.
These effects create an appearance gap far larger than typical natural-image
domain shifts such as day--night or indoor--outdoor variation.
Classical handcrafted feature descriptors (SIFT~\citep{lowe2004sift},
phase congruency) produce too few reliable correspondences under this
gap~\citep{xiang2020automatic}.
Modern deep matchers achieve strong results on ScanNet, MegaDepth, and similar
benchmarks.
Recent cross-modal methods show that modality-aware training can improve
transfer. These include XoFTR for visible--thermal~\citep{tuzcuoglu2024xoftr}
and MatchAnything for multi-domain registration~\citep{han2024matchanything}.
Yet no systematic comparison across the broader matcher ecosystem exists for overhead optical--SAR imagery.

\paragraph{Research question.}
The SpaceNet9 Challenge (conducted summer 2025) formalizes this evaluation with
manually verified tie points across full-resolution optical--SAR scene
pairs~\citep{hansch2024spacenet9,spacenet9page,spacenet9challenge}.
Figure~\ref{fig:hero} previews the operating regime. We estimate tiled
correspondences on large satellite scenes where the optical--SAR modality gap
challenges every stage of the matching pipeline.
A common assumption is that stronger general-purpose matchers transfer broadly.
Our results show that this assumption does not hold uniformly for overhead optical--SAR imagery.
Correspondence quality depends on the matcher architecture, on geometric
post-processing, and on the registration protocol itself.
Tile size, overlap, and inlier gating can change mean error by up to $33\times$ for
a given matcher. Protocol variation can matter more than swapping matchers in
this sweep.

We evaluate twenty-four pretrained matcher configurations under a fixed, zero-shot
geometric protocol, with per-matcher normalization selected on the same three
SpaceNet9 training scenes used for reporting.
The benchmark spans SpaceNet9 and two additional cross-modal datasets
(SRIF~\citep{li2023multimodal}, SARptical~\citep{hughes2018pseudosiamese}).
Rather than proposing a new trained model, we provide controlled evidence of
out-of-domain performance and isolate the failure modes of pretrained matchers.

\paragraph{Contributions.}
\begin{itemize}
  \item A reproducible zero-shot protocol for cross-modal registration
    that tiles large images, filters matches by geometric consistency,
    and measures displacement against verified tie points (\S\ref{sec:method}).
  \item A controlled benchmark of twenty-four pretrained matchers
    showing uneven transfer. Under a fixed geometric protocol with
    per-matcher normalization selected on the labeled training scenes,
    XoFTR and RoMa achieve the lowest reported mean tie-point error ($3.0$\,px) on SpaceNet9.
    MatchAnything-ELoFTR (MA-ELoFTR; $3.4$\,px) is a close third.
    MASt3R/DUSt3R are substantially more protocol-sensitive and less stable
    without protocol tuning (\S\ref{sec:results}, \S\ref{sec:discussion}).
  \item Ablations across 16 protocol configurations (64 total runs) on SpaceNet9
    show that protocol parameters (geometry model, tiling, inlier gating)
    dominate matcher choice as the primary accuracy lever. Findings are
    corroborated on two additional datasets
    (\S\ref{sec:results}).
\end{itemize}

\noindent We define ``good enough'' as sub-8\,px mean tie-point error on
SpaceNet9 under a zero-shot protocol, with rank ordering that is qualitatively
consistent across cross-modal benchmarks.

\section{Related Work}

\paragraph{Cross-modal registration.}
The SpaceNet9 challenge paper frames optical--SAR registration as a
difficult preprocessing problem driven by radiometric and geometric
modality differences, including side-looking SAR distortions and weak
appearance consistency~\citep{hansch2024spacenet9}.
Classical handcrafted pipelines (SIFT~\citep{lowe2004sift}, phase
congruency) are often insufficient under these
conditions~\citep{xiang2020automatic}, motivating deep
feature-matching and learned correspondence methods for SAR--optical
pairs~\citep{hughes2018pseudosiamese,hughes2020deepmatching,zhang2019multimodalfcn}.

\paragraph{Cross-modal matchers and transfer.}
Recent work directly targets cross-modal matching through
modality-aware training.
XoFTR extends LoFTR with masked image modeling pretraining and
pseudo-thermal augmentation for visible--thermal
matching~\citep{tuzcuoglu2024xoftr}, achieving 22\% AUC@5$^\circ$ on
its METU-VisTIR benchmark versus LoFTR's 2.6\%.
However, XoFTR evaluates only on ground-level visible--thermal
pairs; whether this cross-modal design transfers to
the harder overhead optical--SAR domain with speckle, layover, and
orthorectification artifacts has not been tested.
MatchAnything~\citep{han2024matchanything} proposes a large-scale
cross-modality pretraining framework using synthetic
cross-modal pairs and diverse training resources (multi-view
images, video sequences, image warping). Applied to RoMa and ELoFTR base models, their framework improves
SR@10 by 78.5\% and 207.5\% (relative) respectively on a visible--SAR satellite
dataset---the closest prior evidence that cross-modal pretraining can
help overhead SAR registration.
However, MatchAnything evaluates only its own retrained models; no
comparison across the broader matcher ecosystem exists.
RoMa~\citep{edstedt2024roma} takes an orthogonal approach:
frozen DINOv2 features combined with regression-by-classification
matching yield robustness to extreme appearance changes on natural
image benchmarks (36\% relative mean Average Accuracy (mAA) improvement on WxBS), but the method is evaluated exclusively on single-modality perspective imagery.
MINIMA~\citep{jiang2024minima} scales up multimodal training data with
a generative data engine that synthesizes six modalities from RGB
matching pairs; its zero-shot evaluation includes optical--SAR cases,
though only on small patch benchmarks.
Among these cross-modal frameworks, only MatchAnything and MINIMA report
overhead optical--SAR results, in both cases limited to their own
retrained models.
Most image-matching benchmarks use ground-level perspective datasets
(ScanNet, MegaDepth, HPatches), so overhead cross-modal transfer remains
largely unstudied.
This zero-shot transfer gap extends across satellite perception tasks.
Rege Cambrin~\etal~\citep{rege2024depth} show that depth foundation models
require domain-specific adaptation to reliably estimate tree canopy height
from satellite imagery. This result reinforces that satellite data resists
natural-image pretraining assumptions~\citep{rolf2024mission}.
We address this gap by evaluating twenty-four pretrained matchers
\emph{without domain-specific fine-tuning or adaptation}, testing
whether architectural priors and natural-image pretraining alone are
sufficient for cross-modal transfer to overhead optical--SAR imagery.

\paragraph{SpaceNet9 winning solutions.}
Top solutions share several patterns: (i)~patch-wise or tiled
matching with overlap and multi-scale
resizing~\citep{sn9winner2,sn9winner3,sn9winnergrad}; (ii)~local
matches combined with RANSAC and affine or homography estimation,
sometimes with a second refinement
stage~\citep{sn9winner2,sn9winnerug}; (iii)~modality-aware
preprocessing---grayscale conversion, SAR log-compression, CLAHE (Contrast-Limited
Adaptive Histogram Equalization),
black-border handling, building-keypoint
suppression~\citep{sn9winner4,sn9winner5,sn9winnergrad}; and
(iv)~robustness-oriented design over end-to-end training, given
limited labeled scenes~\citep{sn9winner1,sn9winner2,sn9winner3}.
Our work isolates which of these protocol factors matter most by
sweeping them under controlled experimental conditions.

\paragraph{Multimodal benchmarks.}
SRIF provides affine-labeled benchmarks across optical--SAR,
optical--optical, and optical--infrared pairs~\citep{li2023multimodal}.
The concurrent SOMA-1M preprint identifies MapGlue as a strong
matcher for SAR--optical registration~\citep{wu2026soma1m}, but its
data and model artifacts are not yet publicly available in a form
compatible with our evaluation stack, so we treat those results as
related but not directly reproducible evidence.

\section{Method}
\label{sec:method}
\subsection{Task and Data}
\paragraph{Datasets.} We evaluate cross-modal registration on \textit{SpaceNet9 train}, where three
optical--SAR scene pairs include manual tie-point supervision:
Scene~02\_01 (151 tie points, mixed urban/suburban),
Scene~02\_02 (104 tie points, suburban/agricultural), and
Scene~03\_01 (161 tie points, dense urban).
Public test data is reserved for external challenge submission
scoring and is not used for local metric reporting~\citep{spacenet9repo}.

To broaden cross-dataset coverage, we include two additional datasets.
\textbf{SRIF}~\citep{li2023multimodal} contributes 600 valid labeled pairs (200 per modality
split: optical--SAR, optical--optical, optical--infrared), each with
an affine label stored as a $2\times 3$ transform.
\textbf{SARptical}~\citep{hughes2018pseudosiamese} contributes 40 SAR
query patches matched against optical candidate pools (525 total pairs per configuration) for retrieval evaluation.
For each query, we run the matcher against all candidates, fit an affine model via RANSAC (threshold 3.0\,px),
and rank candidates by descending RANSAC inlier count; retrieval metrics (AUROC, AUPRC, Recall@$K$) are computed from this ranking.
Table~\ref{tab:dataset-summary} summarizes all datasets and their
available supervision.

\custompar{Models.}
We evaluate only pretrained models: we apply
no model weight updates (fine-tuning or domain adaptation) on
satellite data. Each tested model has a different training-data
mixture (some include multimodal sources), but as in typical matcher
deployment settings we treat each released checkpoint as a fixed
artifact and measure zero-shot transfer on this cross-modal task.

\begin{table}[t]
  \centering
  \scriptsize
  \setlength{\tabcolsep}{2.5pt}
  \resizebox{0.98\linewidth}{!}{%
    \begin{tabular}{c c c c c}
      \toprule
      Dataset & Modalities & Supervision & Resolution & Coverage \\
      \midrule
      SpaceNet9~\citep{hansch2024spacenet9} & Optical--SAR & Tie
      points & ${\sim}$0.5\,m & 3 scenes \\
      SRIF~\citep{li2023multimodal} & Opt--SAR/Opt--IR & Affine GT
      ($2{\times}3$) & Varies & 600 pairs \\
      SARptical~\citep{hughes2018pseudosiamese} & SAR--Optical
      patches & Binary match/no-match & ${\sim}$1\,m & 40 queries \\
      \bottomrule
    \end{tabular}
  }
  \caption{\textbf{Dataset summary.} Geometry-labeled sets
    (SpaceNet9, SRIF) support transform-grounded evaluation; the
  pair-only set (SARptical) supports matchability and retrieval metrics.}
  \label{tab:dataset-summary}
\end{table}

\subsection{Zero-Shot Matching Protocol}
Our evaluation protocol processes each optical--SAR pair through four stages:

\custompar{1.\ Preprocessing.}
Each image is optionally normalized (identity, percentile clipping to
  $[2, 98]$, z-score, or Contrast-Limited Adaptive Histogram
Equalization (CLAHE) with clip limit~2.0) and resized so the long
side does not exceed a maximum dimension (1024\,px in the primary protocol; varied in the protocol sweep).

\custompar{2.\ Tiled correspondence extraction.}
For large scenes (SpaceNet9), both images are divided into overlapping tiles
($512\times512$\,px tiles with 256\,px overlap in the primary protocol;
tiling, overlap, and resize parameters are varied in the protocol sweep,
\S\ref{sec:results}). Each tile pair is fed to the matcher; tile
coordinates are re-projected to the full-image frame before
aggregation. Tiled inference is motivated by SpaceNet9 winning
solutions that independently converged on similar
strategies~\citep{sn9winner2,sn9winner3}.
SRIF and SARptical pairs are small enough to be processed as single images without tiling.

\custompar{3.\ Geometric filtering.}
Aggregated correspondences (or single-image correspondences for SRIF/SARptical)
are filtered by Random Sample Consensus
(RANSAC)~\citep{fischler1981ransac} with either an affine
($2\times3$) or homography ($3\times3$) model. We use OpenCV's
\texttt{estimateAffine2D} or \texttt{findHomography} with a
reprojection threshold of 3\,px in the primary protocol (swept from
$0.5$ to $20$\,px in \S\ref{sec:extended-ablations}). Pairs producing fewer
inliers than a minimum count (4 in the primary protocol; varied in the ablations)
are marked as failures; for tiled scenes the remaining tiles still contribute.

\custompar{4.\ Displacement prediction.}
The estimated transform is applied to ground-truth tie-point
coordinates (SpaceNet9) or to corner coordinates (SRIF) to produce
predicted displacements, which are compared against ground truth for
metric computation. We adopt fixed random seeds across all runs.
All experiments were run on a single NVIDIA RTX~3090.

\subsection{Metrics}
On SpaceNet9 we report: (i)~\textbf{mean tie-point error} (pixels),
measuring displacement between predicted and ground-truth tie points;
(ii)~\textbf{Success@$\tau$} (S@$\tau$) for $\tau \in \{5,10\}$\,px, the
fraction of tie points with error below $\tau$; and (iii)~\textbf{failure
rate}, the fraction of pairs producing no valid geometric output (due
  to insufficient correspondences, unstable geometry, or degenerate
transforms). Per-pair wall-clock runtime is summarized in
\S\ref{sec:discussion}.
For SRIF, we apply the estimated transform to the four image corners and report
\textbf{mean corner reprojection error} under the dataset-provided affine ground truth;
success metrics use the same thresholds.
For SARptical, where pair-level supervision is available without
geometric labels, we report AUROC, AUPRC, and Recall@$K$.

\subsection{Matchers}
We benchmark twenty-four matcher configurations spanning three architectural
paradigms, all accessed through
\texttt{vismatch}~\citep{vismatch,Berton_2024_EarthMatch}:
\emph{Detector-based:} XFeat and XFeat*~\citep{potje2024xfeat}
(lightweight detector, sparse and semi-dense variants),
ALIKED~\citep{zhao2023aliked}--LightGlue~\citep{lindenberger2023lightglue},
DeDoDe~\citep{edstedt2024dedode}--LightGlue,
SuperPoint~\citep{detone2018superpoint}--LightGlue, and
DISK~\citep{tyszkiewicz2020disk}--LightGlue
(keypoint detectors with learned graph matching),
GIM~\citep{xuelun2024gim}--LightGlue and GIM-DKM (generalist image matchers),
OmniGlue~\citep{jiang2024omniglue} (foundation-model-guided sparse matching),
and SIFT~\citep{lowe2004sift}--LightGlue.
\emph{Detector-free dense:} LoFTR~\citep{sun2021loftr}
(coarse-to-fine transformer), XoFTR~\citep{tuzcuoglu2024xoftr}
(cross-modal LoFTR variant), RoMa and
Tiny-RoMa~\citep{edstedt2024roma}, and RoMaV2~\citep{edstedt2025romav2} (dense warp regression with
DINOv2/3~\citep{oquab2024dinov2, simeoni2025dinov3} backbones), MINIMA-RoMa and
MINIMA-RoMa-Tiny~\citep{jiang2024minima} (multimodal-pretrained RoMa variants),
MINIMA-XoFTR~\citep{jiang2024minima,tuzcuoglu2024xoftr} (multimodal-pretrained XoFTR),
and MA-ELoFTR~\citep{han2024matchanything} (MatchAnything's EfficientLoFTR variant, pretrained on large-scale cross-modal data).
\emph{3D-reconstruction-derived:} MASt3R~\citep{leroy2024mast3r} and
DUSt3R~\citep{wang2024duster}, stereo reconstruction networks
repurposed for 2D matching. The two are used in distinct ways.
MASt3R is used as a \emph{dense feature matcher}: we extract the
per-pixel descriptors produced by its matching head and run fast
reciprocal nearest-neighbor search in descriptor space to obtain 2D
correspondences. DUSt3R is used as a \emph{point-cloud matcher}:
its per-pixel pointmaps are globally aligned, confidence-masked, and
matched by reciprocal nearest neighbors in 3D, with the corresponding
2D pixel indices returned as correspondences. Neither path uses direct
2D-coordinate regression; this distinction matters because the two
failure modes (descriptor ambiguity under speckle for MASt3R vs.\
degenerate pointmaps under near-planar overhead geometry for DUSt3R)
are qualitatively different.
We also include two ensembles (RoMa+LoFTR, RoMa+Tiny-RoMa) that
aggregate correspondences from both matchers before geometric filtering.
We additionally include two classical baselines via
Kornia~\citep{riba2020kornia}: SIFT-NN~\citep{lowe2004sift}, which is
part of the evaluated configurations, and
HardNet~\citep{mishchuk2017hardnet}--LightGlue~\citep{lindenberger2023lightglue},
reported only for SARptical retrieval.
In total, 21~individual matchers, 2~ensembles, and SIFT-NN comprise the
24~evaluated configurations. The classical baselines serve as lower-bound references
with well-understood cross-domain generalization behavior.
We use them to assess whether learned matchers offer meaningful gains over
handcrafted representations on the SAR retrieval task.

\section{Experiments}
\label{sec:experiments}
We run four experiment tracks:
(1)~broad matcher sweeps on SpaceNet9 under deterministic settings;
(2)~matching protocol ablations (geometry model, tiling parameters, inlier gating);
(3)~geometry-labeled cross-dataset evaluation on SRIF; and
(4)~pair-level cross-dataset ranking on SARptical.
All experiments use fixed pair manifests, fixed random seeds, and
identical geometric post-processing per protocol setting.
Runtime is end-to-end per-pair wall-clock on a single RTX~3090, so
speed ranks are comparable within this benchmark.

\custompar{Variance and reproducibility.} All protocols are deterministic (fixed seeds, fixed manifests).
We do not report statistical significance because only three labeled SpaceNet9 scenes are available.
We validate trends on two external datasets, SRIF (600 pairs) and SARptical (40 queries).
This checks whether the same matcher families remain competitive under different datasets and metrics, not whether absolute scores are directly comparable.
All manifests, seeds, and protocol parameters are specified in released configuration files. Code is made available at {\footnotesize{\url{https://github.com/isaaccorley/rsim}}}.

\begin{table}[!t]
  \centering
  \scriptsize
  \resizebox{\columnwidth}{!}{%
    \begin{tabular}{l c r r r r}
      \toprule
      Matcher & Norm & MeanErr$\downarrow$ & S@5$\uparrow$ &
      S@10$\uparrow$ & Fail$\downarrow$ \\
      \midrule
      \multicolumn{6}{l}{\textit{Dense warp (DINOv2/3 backbone)}} \\
      RoMa+Tiny-RoMa~\citep{edstedt2024roma} & Percentile &
      3.6 & 66.6 & 83.8 & 0.00 \\
      RoMaV2~\citep{edstedt2025romav2} & Z-Score &
      3.6 & 69.2 & 90.2 & 0.00 \\
      RoMa+LoFTR~\citep{edstedt2024roma,sun2021loftr} & Percentile &
      3.3 & 75.9 & 90.1 & 0.00 \\
      RoMa~\citep{edstedt2024roma} & Z-Score &
      \textbf{3.0} & \textbf{78.9} & \textbf{94.2} & 0.00 \\
      \midrule
      \multicolumn{6}{l}{\textit{Cross-modal / multimodal-pretrained}} \\
      MINIMA-XoFTR~\citep{jiang2024minima,tuzcuoglu2024xoftr} & Percentile &
      3.8 & 64.6 & 77.3 & 0.00 \\
      MINIMA-RoMa~\citep{jiang2024minima,edstedt2024roma} & Percentile &
      3.4 & 66.6 & 79.6 & 0.00 \\
      MA-ELoFTR~\citep{han2024matchanything} & Z-Score &
      3.4 & 71.2 & 93.3 & 0.00 \\
      XoFTR~\citep{tuzcuoglu2024xoftr} & Percentile &
      \textbf{3.0} & 78.4 & 90.5 & 0.00 \\
      \midrule
      \multicolumn{6}{l}{\textit{Detector + graph matcher (LightGlue)}} \\
      DISK-LG~\citep{tyszkiewicz2020disk,lindenberger2023lightglue} & Z-Score &
      4.6 & 52.2 & 82.6 & 0.67 \\
      ALIKED-LG~\citep{zhao2023aliked,lindenberger2023lightglue} & Percentile &
      4.5 & 62.2 & 84.8 & 0.00 \\
      DeDoDe-LG~\citep{edstedt2024dedode,lindenberger2023lightglue} & Percentile &
      4.5 & 58.8 & 85.5 & 0.00 \\
      SuperPoint-LG~\citep{detone2018superpoint,lindenberger2023lightglue} & Z-Score &
      4.4 & 65.4 & 77.6 & 0.00 \\
      GIM-DKM~\citep{xuelun2024gim} & Z-Score &
      4.1 & 57.6 & 85.6 & 0.00 \\
      \midrule
      \multicolumn{6}{l}{\textit{Detector-free dense}} \\
      LoFTR~\citep{sun2021loftr} & Identity &
      5.1 & 54.4 & 81.0 & 0.00 \\
      \midrule
      \multicolumn{6}{l}{\textit{3D-reconstruction-derived}} \\
      MASt3R~\citep{leroy2024mast3r} & Identity &
      4.5 & 60.9 & 80.9 & 0.00 \\
      \bottomrule
    \end{tabular}
  }
  \caption{\textbf{SpaceNet9: Top-15 matchers (best norm per matcher).}
    All 24~matchers evaluated under a single protocol:
    tiled inference (512\,px tiles, 256\,px overlap), affine geometry,
    1024\,px max side, RANSAC threshold 3.0\,px, min 4~inliers.
    Each row shows the best normalization for that matcher.
    MeanErr in pixels; S@$k$ in \%; Fail in [0,1].}
  \label{tab:final-master-ranking}
\end{table}

\section{Results and Analysis}
\label{sec:results}

\subsection{SpaceNet9 Results}
Table~\ref{tab:final-master-ranking} ranks the top 15~matchers on SpaceNet9 under a fixed geometric protocol (affine geometry, 512\,px tiles, 256\,px overlap, 1024\,px max side, RANSAC threshold 3.0\,px, $\geq4$~inliers). Each row reports the best normalization for that matcher, selected on the same three
labeled training scenes used for reporting; all other protocol variables are held constant, enabling
direct cross-matcher comparison. Configurations with persistent full failure are excluded from the ranking.
These results describe performance on the labeled training scenes, not generalization estimates.
We assess ordering stability on SRIF and SARptical below.

\begin{table}[ht!]
  \centering
  \scriptsize
  \resizebox{\columnwidth}{!}{\begin{tabular}{lccccc}
\toprule
Matcher & Norm & MeanErr (px)$\downarrow$ & S@5$\uparrow$ & S@10$\uparrow$ & Fail$\downarrow$ \\
\midrule
SuperPoint-LightGlue~\cite{lindenberger2023lightglue} & CLAHE & 67.0 & 31.3 & 39.0 & 0.64 \\
RoMa~\cite{edstedt2024roma} & CLAHE & 64.9 & 36.0 & 40.4 & \textbf{0.00} \\
RoMa+LoFTR~\cite{edstedt2024roma} & CLAHE & 64.6 & 36.1 & 40.6 & \textbf{0.00} \\
RoMa+Tiny-RoMa~\cite{edstedt2024roma} & CLAHE & 63.9 & 35.3 & 40.0 & \textbf{0.00} \\
LoFTR~\cite{sun2021loftr} & CLAHE & 63.3 & 29.7 & 39.4 & 0.65 \\
LoFTR~\cite{sun2021loftr} & Z-Score & 63.1 & 30.6 & 41.7 & 0.64 \\
LoFTR~\cite{sun2021loftr} & Identity & 59.3 & 31.9 & 42.3 & 0.66 \\
MINIMA-RoMa~\cite{jiang2024minima} & Identity & 48.2 & \textbf{42.0} & \textbf{45.8} & \textbf{0.00} \\
MINIMA-RoMa~\cite{jiang2024minima} & CLAHE & 47.7 & 40.4 & 44.7 & \textbf{0.00} \\
MINIMA-RoMa~\cite{jiang2024minima} & Z-Score & \textbf{47.0} & 41.7 & 45.7 & \textbf{0.00} \\
\bottomrule
\end{tabular}}
  \caption{\textbf{SRIF: Normalization ablation (600~pairs).}
    Top matchers from SpaceNet9 evaluated on SRIF across all three
  normalizations, showing sensitivity to preprocessing choice.}
  \label{tab:srif-ablation-results}
\end{table}

\begin{table}[ht!]
  \centering
  \scriptsize
  \resizebox{\columnwidth}{!}{\begin{tabular}{lccccc}
\toprule
Matcher & Norm & AUROC$\uparrow$ & AUPRC$\uparrow$ & R@1$\uparrow$ & R@5$\uparrow$ \\
\midrule
Tiny-RoMa~\cite{edstedt2024roma} & CLAHE & 0.52 & 0.10 & 7.5 & 35.0 \\
SuperPoint-LightGlue~\cite{lindenberger2023lightglue} & Z-Score & 0.53 & 0.09 & 67.5 & \textbf{100.0} \\
DISK-LightGlue~\cite{lindenberger2023lightglue} & Identity & 0.53 & 0.09 & 70.0 & 97.5 \\
LoFTR~\cite{sun2021loftr} & Identity & 0.54 & 0.10 & 10.0 & 50.0 \\
HardNet-LightGlue~\cite{mishchuk2017hardnet} & Identity & 0.54 & 0.11 & 25.0 & 50.0 \\
DUSt3R~\cite{wang2024duster} & CLAHE & 0.55 & 0.10 & 92.5 & \textbf{100.0} \\
SIFT-NN~\cite{lowe2004sift} & Identity & 0.55 & 0.11 & 92.5 & \textbf{100.0} \\
XoFTR~\cite{tuzcuoglu2024xoftr} & Percentile & 0.55 & 0.09 & 2.5 & 57.5 \\
MA-ELoFTR~\cite{han2024matchanything} & Z-Score & 0.56 & 0.11 & \textbf{97.5} & \textbf{100.0} \\
MINIMA-RoMa~\cite{jiang2024minima} & Z-Score & \textbf{0.57} & \textbf{0.11} & 12.5 & 52.5 \\
\bottomrule
\end{tabular}}
  \caption{\textbf{SARptical: Top-10 configurations by AUROC.} Each
    configuration uses 40~SAR queries over 525~candidate pairs;
    AUROC/AUPRC measure pair-level ranking quality over the full candidate pool, while
  Recall@$K$ measures shortlist hit rate; these metrics can diverge.}
  \label{tab:sarptical-grid}
\end{table}

\subsection{Protocol Sweep}
\begin{figure}[tbp]
  \centering
  \includegraphics[width=0.90\linewidth]{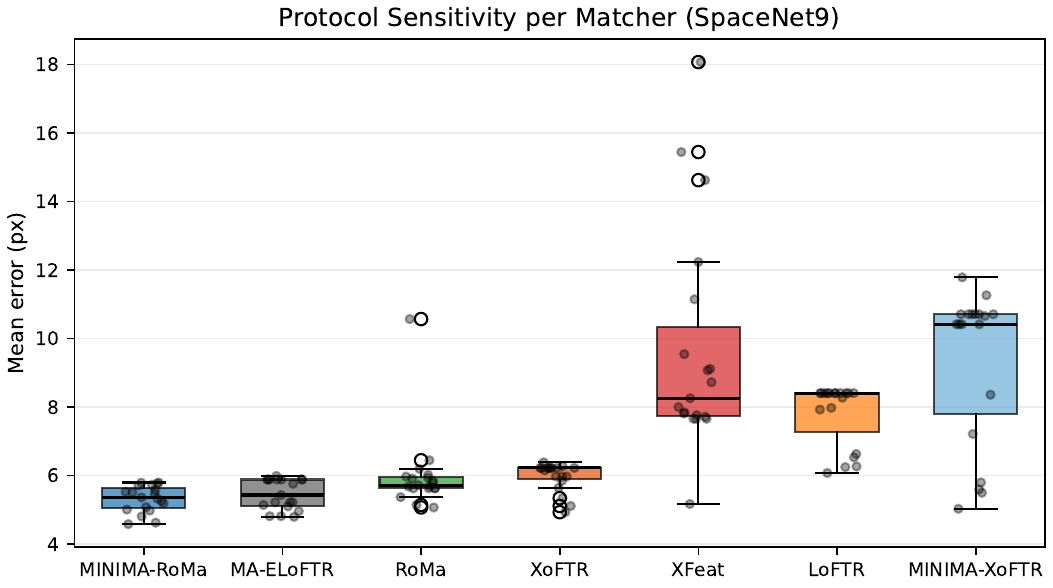}
  \caption{\textbf{Protocol sensitivity per matcher (SpaceNet9).}
    Y-axis: mean error (px, $\downarrow$ better). Each data point is one result from the
    threshold-robustness, keypoint-budget, or inlier-gating ablations, aggregated across seven
    matchers (MA-ELoFTR, MINIMA-RoMa, RoMa, XoFTR, LoFTR, MINIMA-XoFTR, XFeat; 133 total runs).
    \textit{Intra}-matcher variance often matches or exceeds \textit{inter}-matcher differences.
    Protocol choices can matter as much as matcher selection in the evaluated ablations.}
  \label{fig:protocol-sensitivity-violin}
\end{figure}

\paragraph{Protocol sensitivity can exceed matcher differences.}
Figure~\ref{fig:protocol-sensitivity-violin} aggregates the threshold,
keypoint-budget, and inlier-gating ablations across seven matchers.
Intra-matcher error variance
across protocol configurations is often comparable to or larger than inter-matcher differences,
suggesting that protocol tuning is at least as important as matcher selection on SpaceNet9.
Even XFeat, whose sparser keypoint set requires permissive RANSAC
thresholds ($t{\geq}5$\,px) to reach competitive inlier counts, gains a
$>$2$\times$ error reduction from protocol tuning.

We sweep 16~protocol configurations---2~geometry models (affine, homography) $\times$
2~max resize values (1024, 1536\,px) $\times$ 2~tile overlaps (128, 256\,px) $\times$
2~minimum inlier counts (4, 8)---across 2~matchers (LoFTR, XFeat) and 2~normalizations
(identity, percentile), yielding 64~total runs on all labeled SpaceNet9 train pairs (three scenes).
Tile size is fixed at 512\,px and the RANSAC reprojection threshold at 3.0\,px throughout.
Keypoint budget is fixed at 4096 for all sweep runs; its effect is ablated separately in \S\ref{sec:extended-ablations}.

Two trends are consistent:
(1)~affine geometry outperforms homography ($9.7$ vs.\ $12.3$\,px mean error; $0.707$ vs.\ $0.658$ S@10).
This is expected for orthorectified satellite imagery: orthorectification removes
perspective distortion, making the true sensor-to-sensor transform approximately
affine (translation, rotation, scale, and shear).
The additional perspective degrees of freedom in a homography ($3\times3$, 8~DoF)
may therefore absorb noise rather than signal, consistent with the degraded
accuracy we observe relative to the 6-parameter affine model.
(2)~LoFTR with percentile normalization under affine geometry yields
the best configuration ($7.2$\,px mean error, $0.802$ S@10, 0\% failure).
Given the three-scene label coverage, we do not claim quantitative
significance.

\begin{figure*}[tbp]
  \centering
  \begin{subfigure}[t]{0.33\textwidth}
    \centering
    \includegraphics[width=\linewidth]{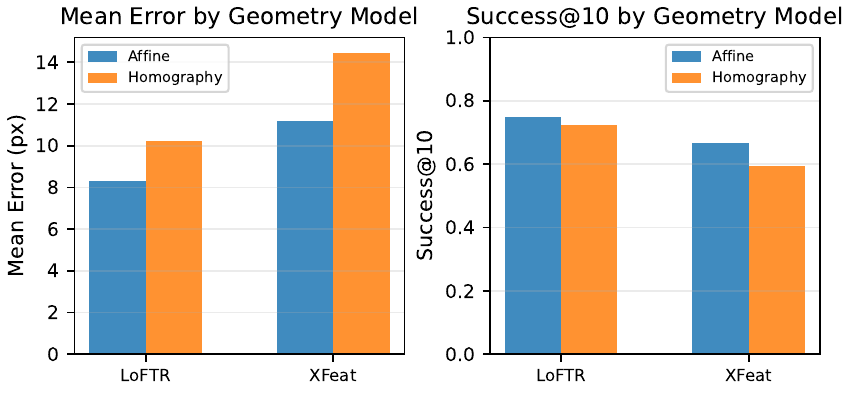}
    \caption{\textbf{Geometry-model comparison on SpaceNet9.} Affine
      consistently improves both mean error ($\downarrow$) and S@10 ($\uparrow$)
      over homography under the evaluated sweep configurations.
      Orthorectification makes the true transform approximately affine,
    so homography's extra degrees of freedom likely absorb noise rather than signal.}
    \label{fig:spacenet-geometry-tradeoff}
  \end{subfigure}\hspace{0.03\textwidth}
  \begin{subfigure}[t]{0.58\textwidth}
    \centering
    \includegraphics[width=\linewidth]{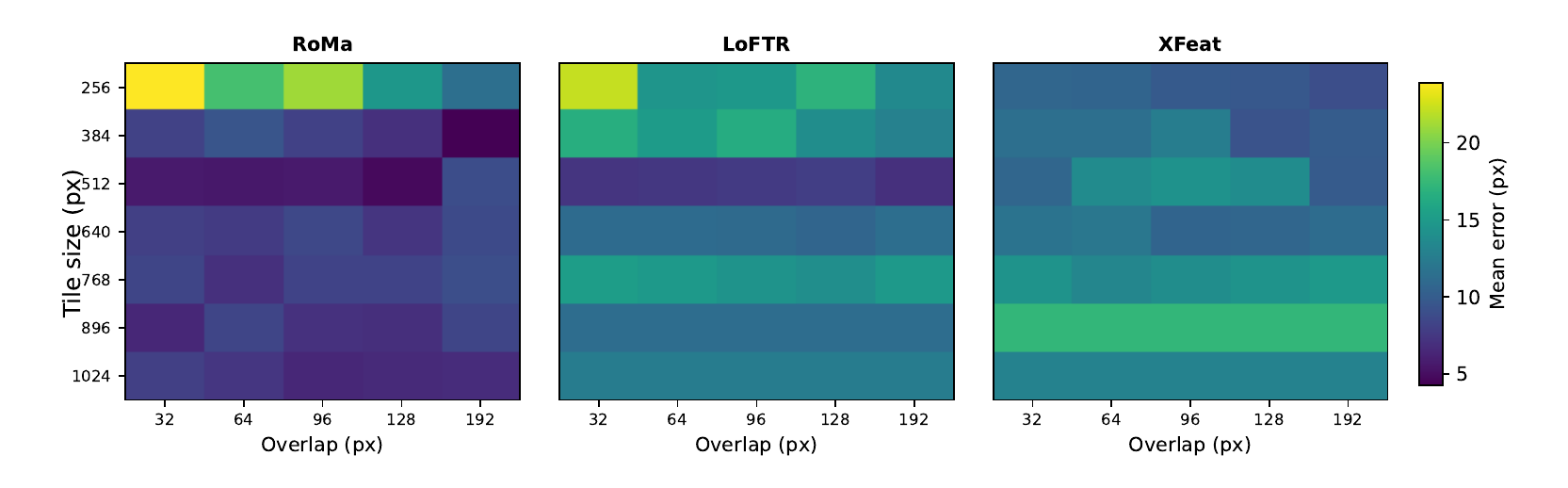}
    \caption{\textbf{Tile size vs.\ overlap heatmaps
      (RoMa/LoFTR/XFeat).} Mean error (px) under a fixed affine tiled
    protocol. Generally, the tile size is more influential than overlap, with results favoring tile sizes near each matcher's training resolution.}
    \label{fig:spacenet-tile-overlap-heatmap}
  \end{subfigure}
  \caption{\textbf{Core protocol effects on SpaceNet9.} Geometry
  choice and tiling interact strongly with matcher performance.}
\end{figure*}

Figure~\ref{fig:spacenet-geometry-tradeoff} visualizes the affine
vs.\ homography gap.
Figure~\ref{fig:spacenet-tile-overlap-heatmap} shows the joint effect
of tile size and overlap on accuracy for RoMa, LoFTR, and XFeat under
a fixed protocol: too-small tiles with low overlap are unstable
while moderate overlap improves robustness.
Per-normalization sensitivity is captured in the SRIF heatmap
(Figure~\ref{fig:norm-matcher-heatmap}).

\subsection{Extended Transfer Ablations}
\label{sec:extended-ablations}
We run threshold and inlier-gating ablations on the top SpaceNet9 matchers:
\emph{threshold robustness} sweeps $t \in \{0.5, 1, 2, 3, 5, 8, 10, 15, 20\}$\,px
across eight matchers (MA-ELoFTR, MINIMA-RoMa, RoMa, RoMaV2, XoFTR, LoFTR,
MINIMA-XoFTR, XFeat), and
\emph{inlier-gating sensitivity} sweeps minimum inlier count $\in \{2, 4, 6, 8, 10\}$.
The best-performing configuration, MA-ELoFTR with threshold $t{=}15$\,px, achieves
S@10${=}0.948$ and $4.8$\,px mean error; MINIMA-RoMa follows at S@10${=}0.901$.
Inlier gating is stable across all tested settings.

The RANSAC threshold sweep reveals non-monotonic behavior.
Across matcher families, tighter thresholds improve geometric
precision for strong dense matchers, while weaker or sparser matchers
often require more permissive thresholds to recover enough inliers.
This heterogeneity is consistent with Edstedt's finding that fixed
RANSAC thresholds are suboptimal across
domains~\citep{edstedt2025thresholdrobust}.
Inlier-gating ablations show comparatively stable behavior once tiled
affine fitting is active, indicating that correspondence quality and
threshold selection dominate over minimum inlier count in this regime.

\begin{figure}[tbp]
  \centering
  \includegraphics[width=0.90\linewidth]{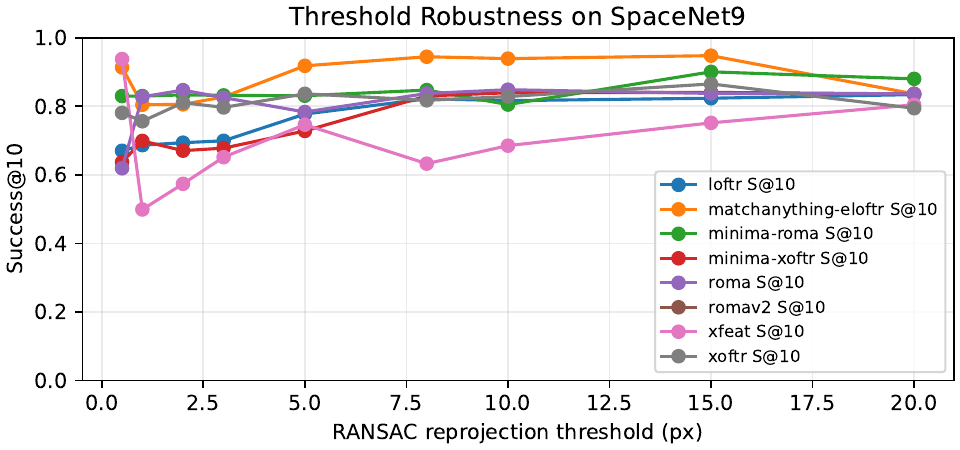}
  \caption{\textbf{RANSAC threshold robustness.} S@10 ($\uparrow$ better)
    as a function of reprojection threshold (px) for eight matchers under fixed
    tiled affine settings. MA-ELoFTR and MINIMA-RoMa maintain
    $>$90\% success at permissive thresholds; XoFTR and RoMa follow
    closely. Sparser or less robust matchers (LoFTR, XFeat) are more
    threshold-sensitive. Their deployment accuracy depends strongly on threshold selection.}
  \label{fig:spacenet-threshold-robustness}
\end{figure}

\paragraph{SARptical expanded evaluation.}
We evaluate all SpaceNet9-selected matchers plus SIFT, DISK, and HardNet-LightGlue on 40~SAR queries with
525~candidate pairs per configuration.
Table~\ref{tab:sarptical-grid} summarizes results.

MINIMA-RoMa leads pair-ranking quality under this protocol (AUROC range across all configurations: 0.52--0.57).
The AUROC margins are narrow, which suggests some cross-dataset robustness.
Among classical baselines, SIFT-NN is competitive but trails MINIMA-RoMa,
while HardNet-LightGlue ranks near chance---its descriptor space does not generalize to SAR imagery.
Top-10 configurations by AUROC are in Table~\ref{tab:sarptical-grid}.

\custompar{Retrieval.}
Recall@$K$ rises sharply with shortlist depth, reaching $1.0$ by $K{=}10$,
yet high Recall@$K$ does not imply strong global ranking (AUROC/AUPRC)---top-1
ranking remains hard under SAR--optical shift.

Across all ablations, two factors dominate accuracy in our sweeps:
(i)~the RANSAC reprojection threshold and
(ii)~the keypoint budget; retrieval shortlist depth matters for SARptical ranking,
while minimum inlier count is comparatively insensitive once tiled affine fitting is active.
Given this sensitivity profile, improving geometric verification (e.g.,
threshold-robust RANSAC) and multi-stage retrieval pipelines (e.g., coarse
candidate selection followed by spatial verification re-ranking) may yield
gains comparable to or larger than switching matchers.

\subsection{Cross-Dataset Transfer (SRIF)}
To test whether SpaceNet9 matcher rankings generalize to other datasets and modality
pairs, we evaluate on SRIF (affine supervision, 600~pairs) using the same deterministic
protocol family. This complements the SARptical retrieval evidence above.
Because SpaceNet9, SRIF, and SARptical use different primary metrics (mean tie-point
error, mean corner reprojection error, and AUROC respectively), computed at different
image resolutions and scene scales, absolute errors are not comparable across
datasets (a matcher's SRIF error being an order of magnitude above its SpaceNet9
error reflects the metric and dataset, not a performance drop); only relative
rank ordering transfers.
When comparing SRIF results, we report both the lowest mean corner reprojection error and the best zero-failure
trade-off, since several methods reduce error at the cost of substantial failure rates.
Table~\ref{tab:srif-ablation-results} reports rankings over the
SpaceNet9-selected matcher pool.

\begin{figure}[ht!]
  \centering
  \includegraphics[width=0.95\linewidth]{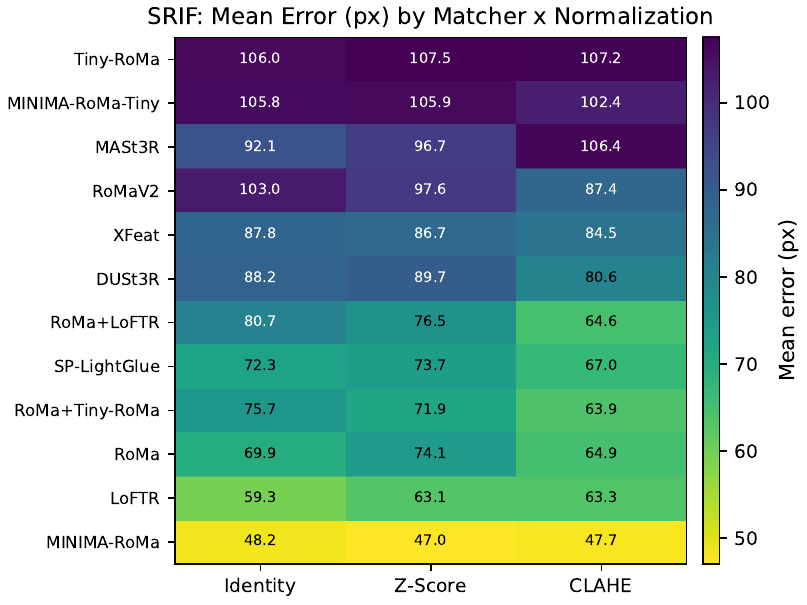}
  \caption{\textbf{Normalization $\times$ matcher heatmap (SRIF).} Cell
    values are mean corner reprojection error (px); greater column-wise
    spread indicates greater sensitivity to the normalization scheme.
    MINIMA-RoMa with Z-Score achieves the lowest error.}
  \label{fig:norm-matcher-heatmap}
\end{figure}

MA-ELoFTR achieves the lowest overall mean corner reprojection error on SRIF ($45.6$\,px, best observed under this protocol),
though at a higher failure rate ($0.46$).
Among zero-failure methods on SRIF, MINIMA-RoMa ($47.0$\,px, $0\%$ failure)
shows the best accuracy--reliability trade-off.
This pattern is consistent with improved cross-dataset robustness for MINIMA-RoMa,
especially among zero-failure methods, though the datasets and primary metrics differ.
Combined with the SARptical results above, MINIMA-RoMa shows the strongest zero-failure cross-dataset consistency among the evaluated methods, even though the SpaceNet9 top rank is shared by RoMa/XoFTR\@.
On SRIF's easier splits (e.g., optical--optical),
sparse detectors perform well because appearance differences are small;
on harder modality gaps (optical--SAR, optical--infrared), detector-free dense matchers hold the advantage.
Figure~\ref{fig:norm-matcher-heatmap} visualizes the
normalization$\times$matcher interaction on SRIF: MINIMA-RoMa with
Z-Score normalization achieves the lowest error, but the column-wise
spread indicates that normalization sensitivity varies substantially across architectures.

\subsection{Discussion}
\label{sec:discussion}

\begin{figure*}[tbp]
  \centering
  \setlength{\tabcolsep}{1pt}
  \renewcommand{\arraystretch}{0.7}
  \resizebox{0.85\textwidth}{!}{%
  \begin{tabular}{@{}ccc@{}}
    \footnotesize\textbf{RoMa / Z-Score} &
    \footnotesize\textbf{LoFTR / CLAHE} &
    \footnotesize\textbf{XFeat / Identity} \\
    \includegraphics[width=0.333\textwidth]{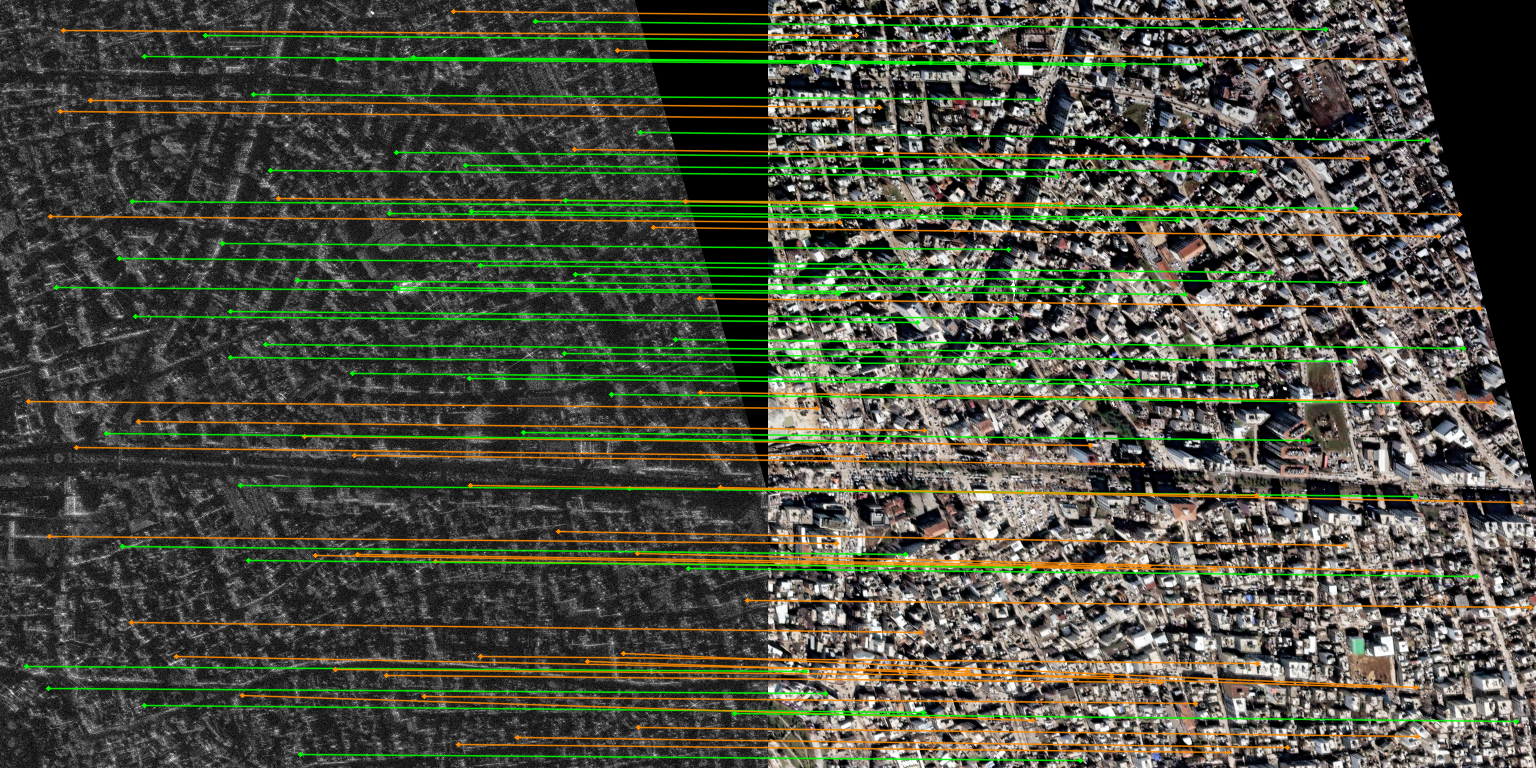} &
    \includegraphics[width=0.333\textwidth]{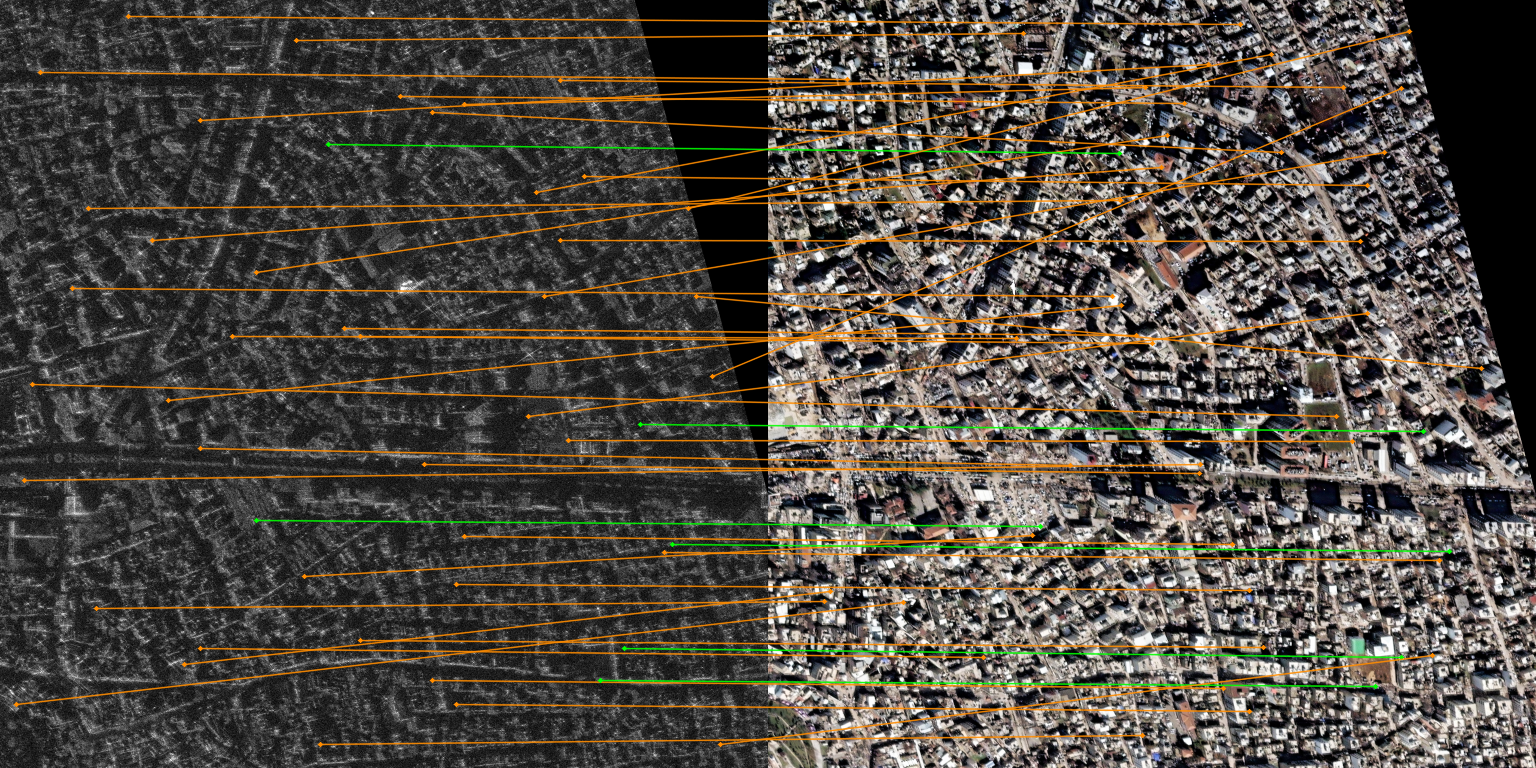} &
    \includegraphics[width=0.333\textwidth]{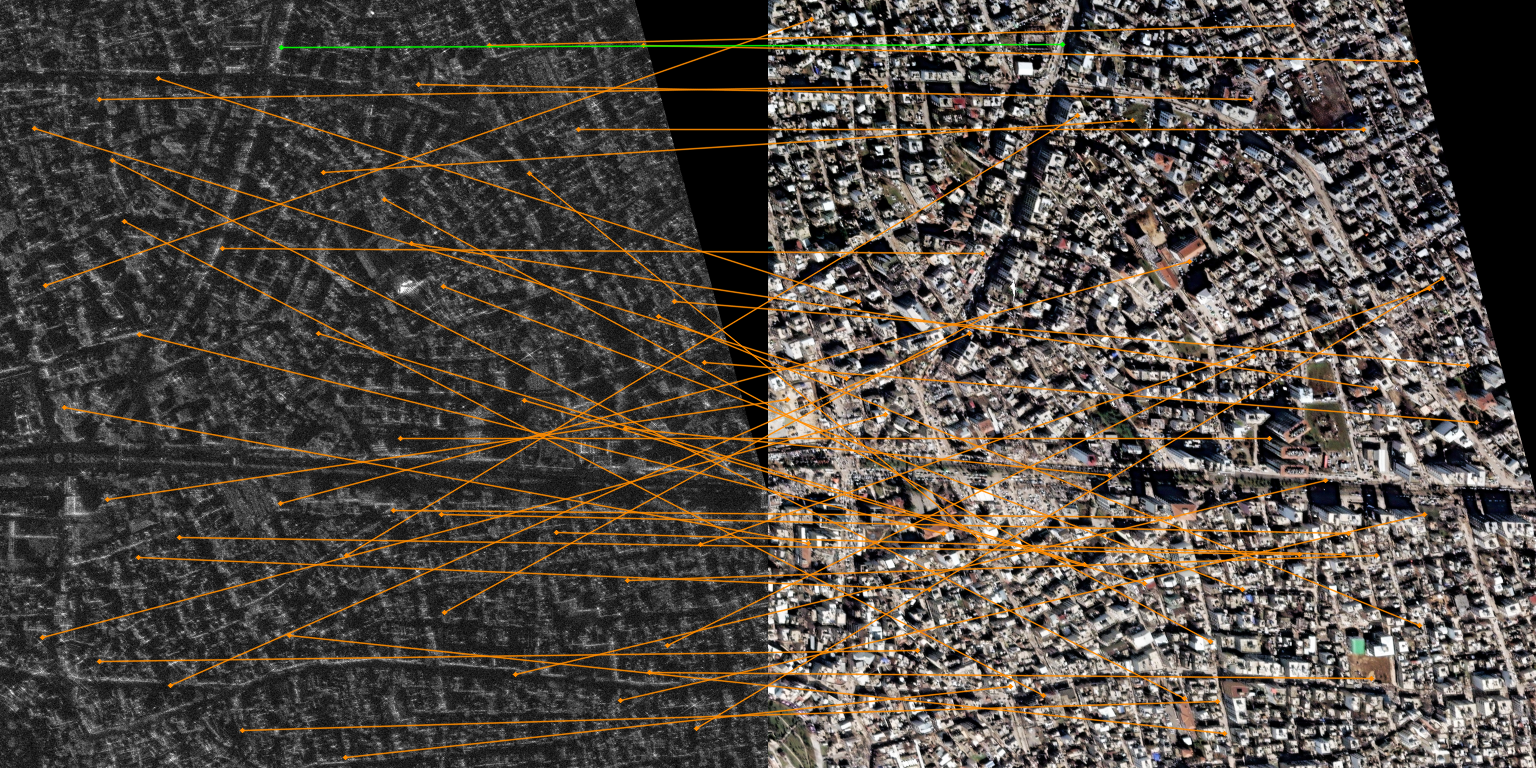} \\[0pt]
    \footnotesize\textbf{SuperPoint-LG / Z-Score} &
    \footnotesize\textbf{Tiny-RoMa / CLAHE} &
    \footnotesize\textbf{RoMa+Tiny-RoMa / Identity} \\
    \includegraphics[width=0.333\textwidth]{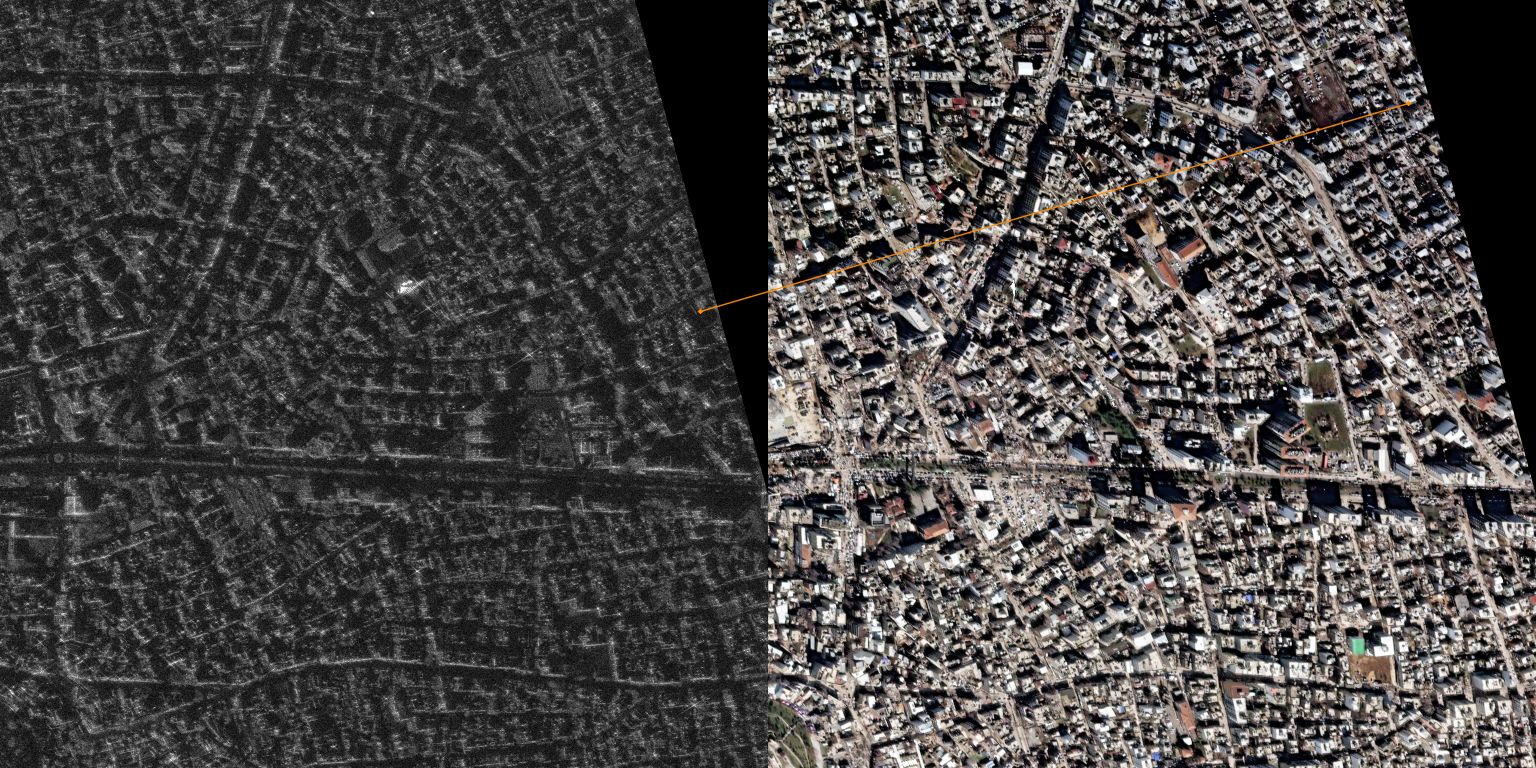} &
    \includegraphics[width=0.333\textwidth]{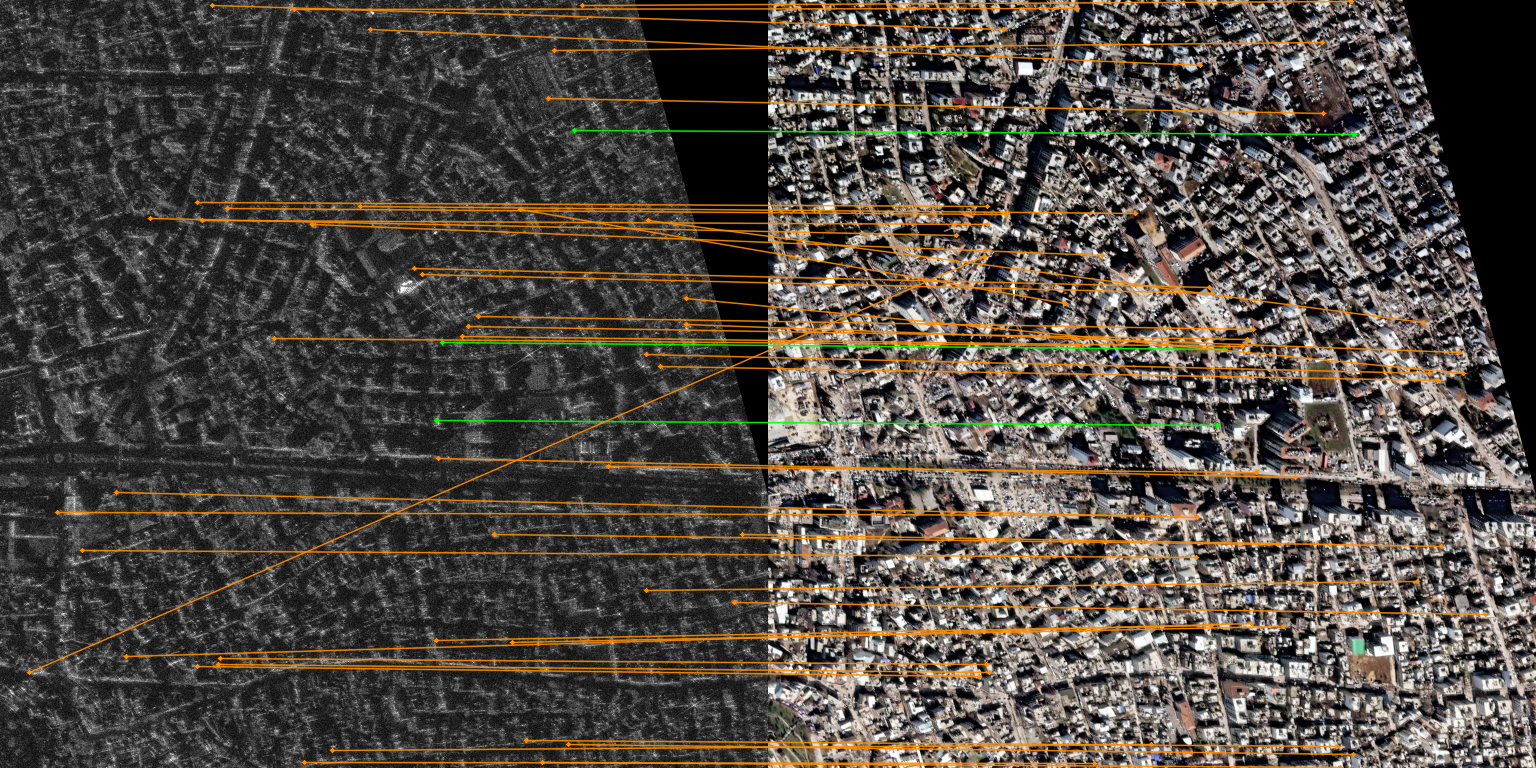} &
    \includegraphics[width=0.333\textwidth]{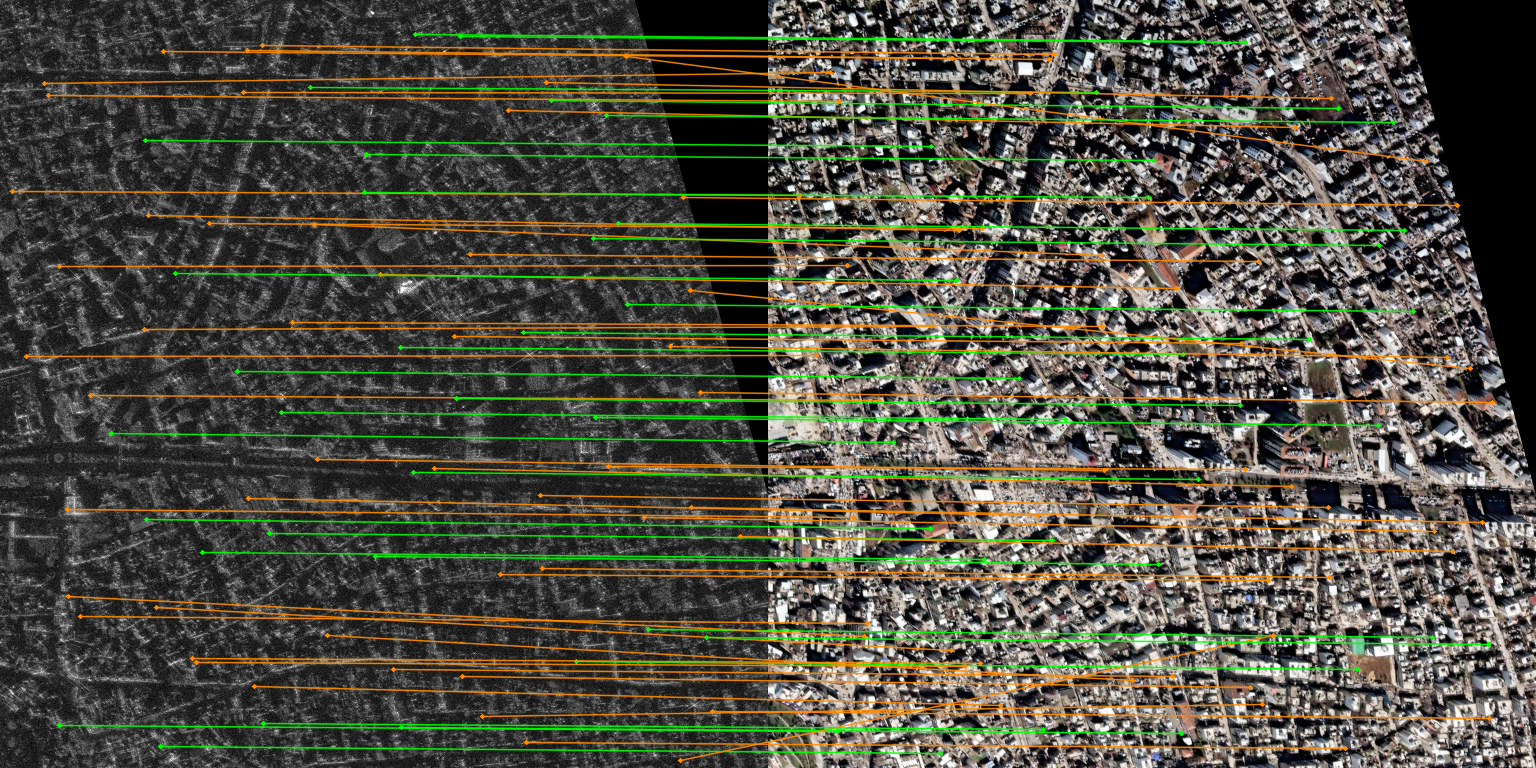} \\[0pt]
    \footnotesize\textbf{RoMa+LoFTR / CLAHE} &
    \footnotesize\textbf{MASt3R / CLAHE} &
    \footnotesize\textbf{DUSt3R / CLAHE} \\
    \includegraphics[width=0.333\textwidth]{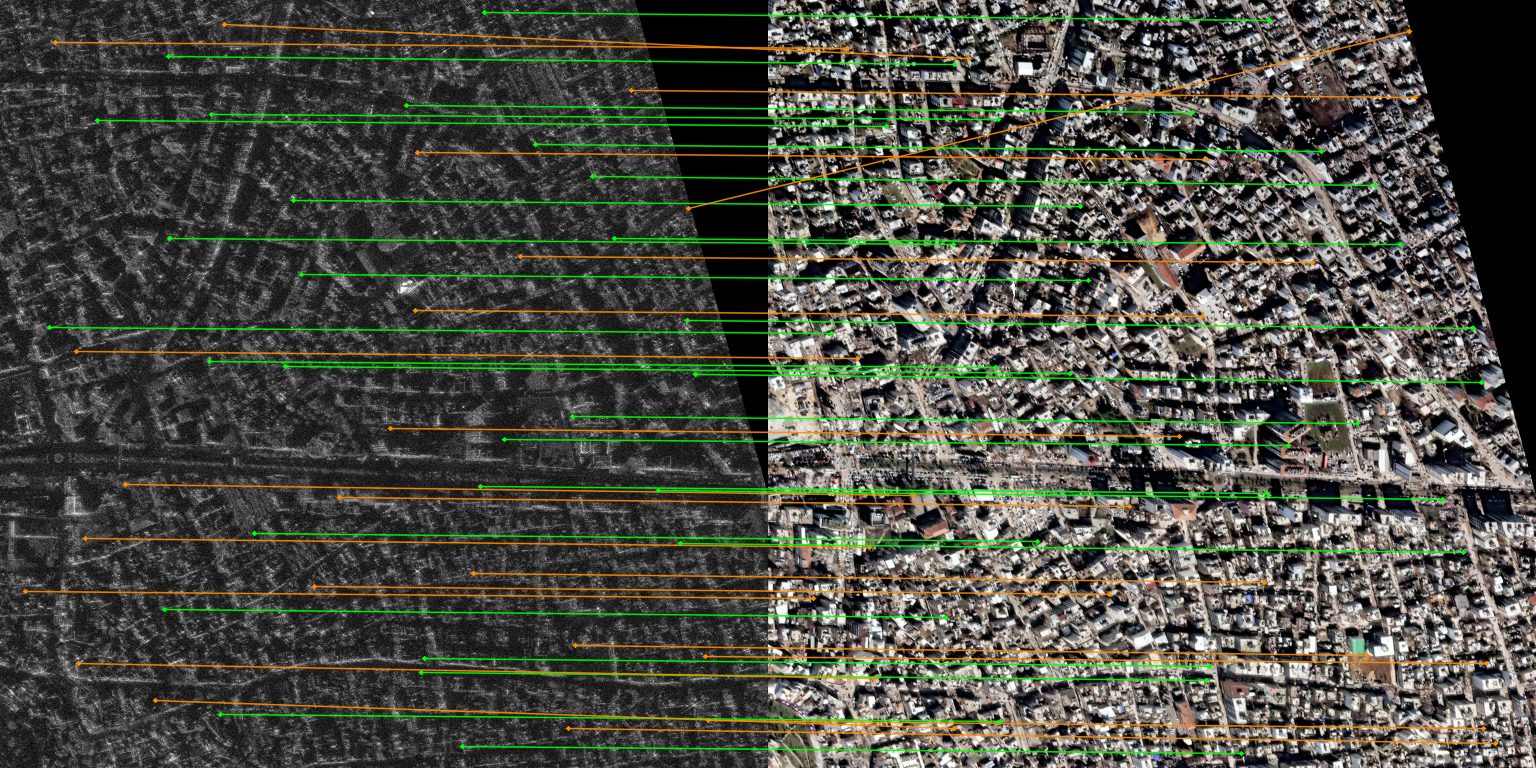} &
    \includegraphics[width=0.333\textwidth]{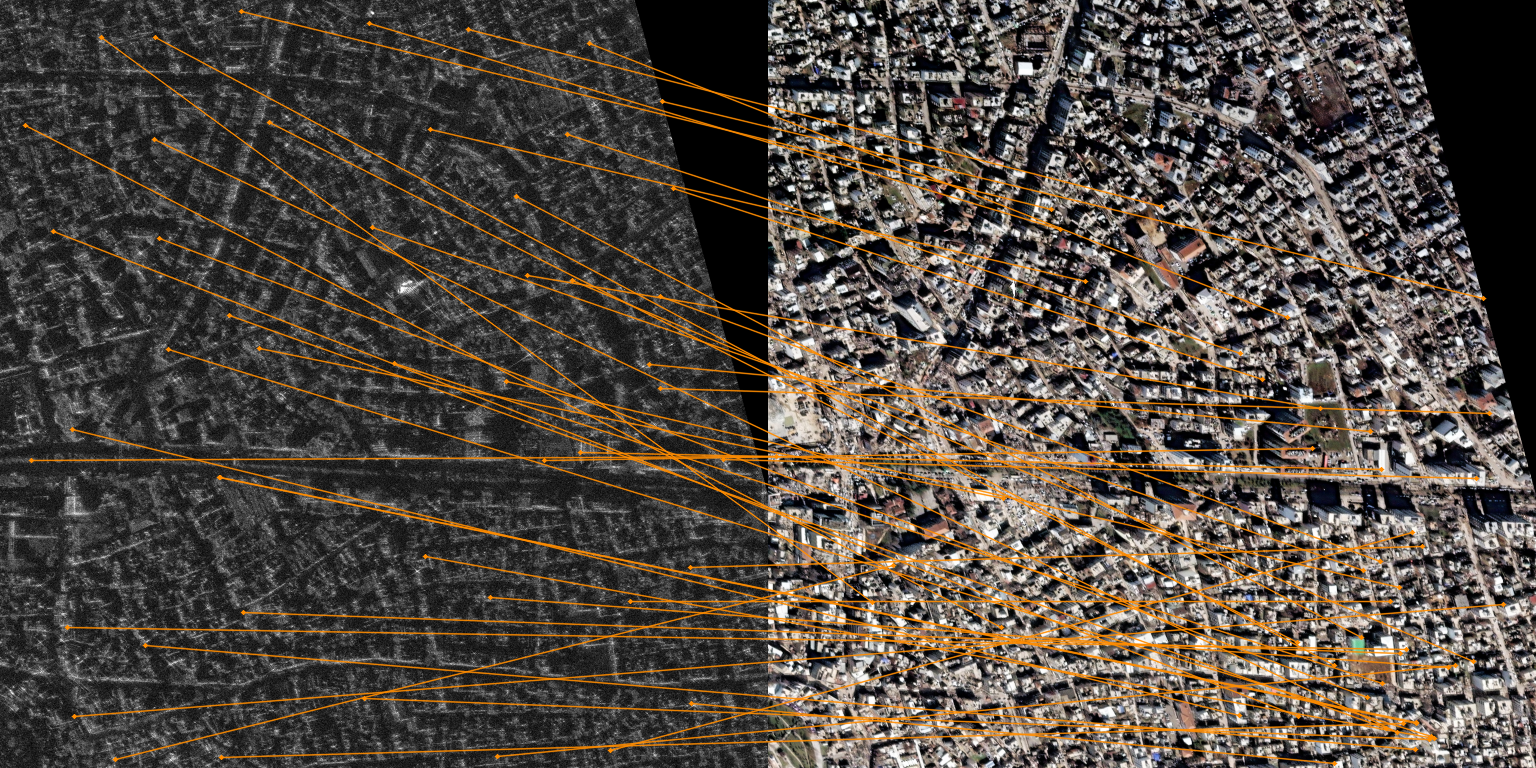} &
    \includegraphics[width=0.333\textwidth]{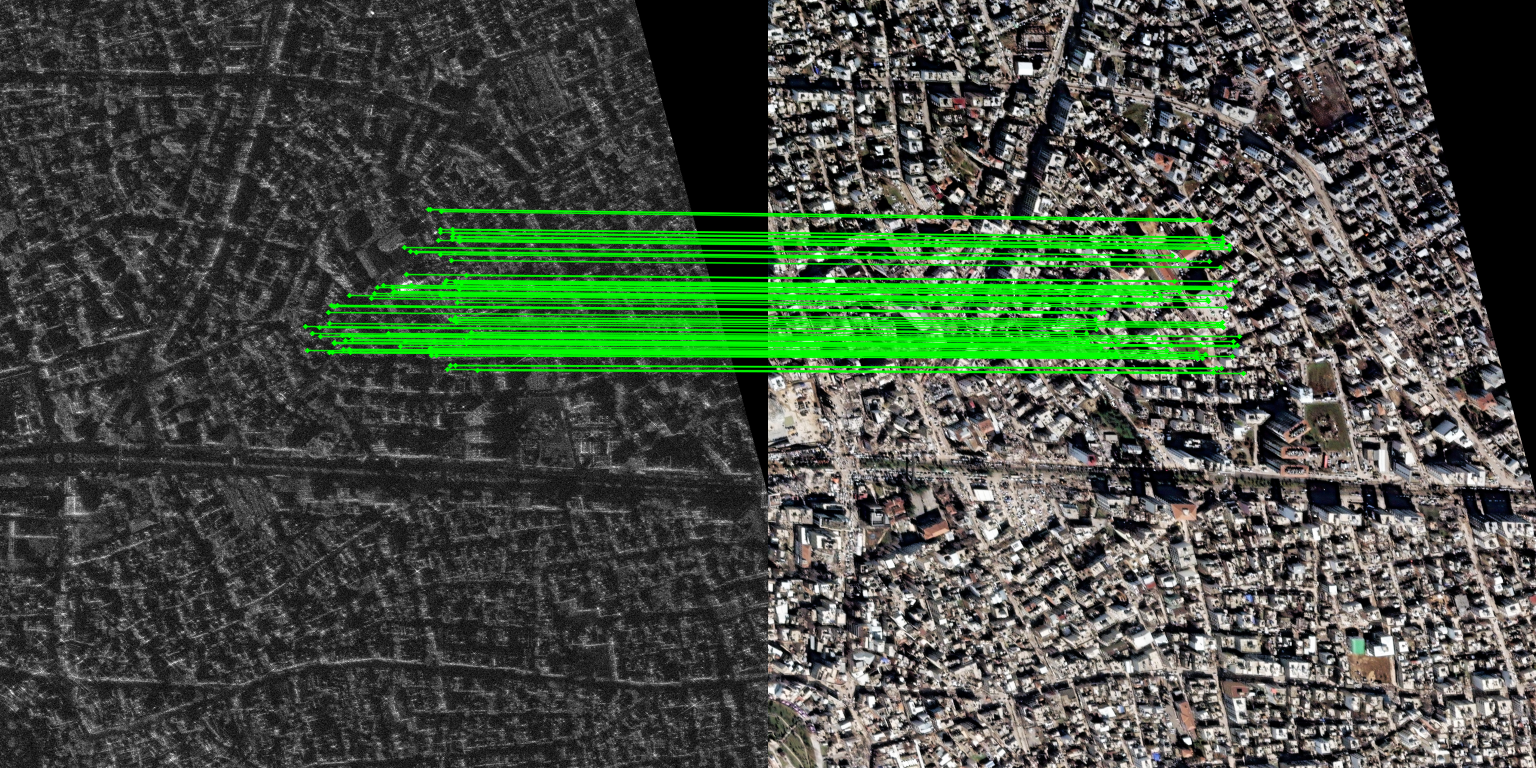} \\
  \end{tabular}
  }
  \caption{\textbf{SpaceNet9 qualitative correspondence gallery.}
    Panels show top matchers under their best zero-shot configurations; green lines denote affine-RANSAC inliers and orange lines matched outliers.}
  \label{fig:spacenet9-correspondence-gallery}
\end{figure*}

\paragraph{Why do some matchers transfer and others fail?}
XoFTR's strong performance (lowest reported mean tie-point error on SpaceNet9, tied with RoMa) is consistent
with its design. Masked image modeling pretraining and pseudo-thermal
augmentation explicitly target cross-modal
invariance~\citep{tuzcuoglu2024xoftr}. Transfer to optical--SAR is therefore
expected, even though XoFTR was trained only on visible--thermal ground-level
imagery.
More surprising is RoMa~\citep{edstedt2024roma}, which is trained
exclusively on single-modality perspective imagery (MegaDepth) yet ties
XoFTR's $3.0$\,px mean tie-point error on SpaceNet9, suggesting that its
architecture may provide partial modality invariance.
MINIMA-RoMa~\citep{jiang2024minima}, which does include multimodal
pretraining on synthetic cross-modal data, also ranks near the top.
RoMa uses dense warp regression over frozen DINOv2~\citep{oquab2024dinov2}
features.
DINOv2 is not explicitly trained for modality invariance; its web-scale pretraining
may capture cross-modal correlations sufficient for partial SAR transfer.
This is a plausible explanation, not a causal claim.
MASt3R and DUSt3R, designed for 3D pointmap regression rather than
2D matching, assume perspective geometry and scene depth variation, both of
which are violated by orthorectified satellite imagery.
As described in \S\ref{sec:method}, MASt3R is repurposed as a dense
feature matcher via its descriptor head and DUSt3R as a point-cloud
matcher via reciprocal 3D-NN over its pointmaps, so their failure
modes are not identical. DUSt3R's pointmap path is most severely affected
by near-planar overhead geometry: its 3D regression heads produce
geometrically inconsistent pointmaps when depth variation is minimal,
yielding degenerate reciprocal matches that cause RANSAC to fail.
MASt3R's descriptor path appears to degrade more gracefully under the
same geometry (consistent with its top-15 placement in
Table~\ref{tab:final-master-ranking}), though it likely still suffers
from weak descriptor discriminability under speckle and radiometric
inversion. Their training mixtures (eight datasets for DUSt3R,
e.g., ScanNet++, CO3D-v2, MegaDepth, and Waymo, extended further for
MASt3R) span indoor, object-centric, photogrammetric, and driving
scenes, but include no satellite or SAR imagery, unlike DINOv2's
web-scale pretraining. This geometric and domain mismatch explains their
protocol sensitivity.

\custompar{Protocol vs.\ matcher: which matters more?}
As quantified in Figure~\ref{fig:protocol-sensitivity-violin} and discussed in \S\ref{sec:results},
protocol choices can equal or exceed matcher selection effects in the evaluated configurations.
Relaxing MASt3R's tiling protocol yields a $33\times$ error reduction, larger than the gap between the worst and best matchers under any fixed protocol. Similarly, switching from homography to affine geometry reduces mean
error by $2.6$\,px on average (Figure~\ref{fig:spacenet-geometry-tradeoff}). This gap is comparable to the
difference between mid-tier and top-tier matchers and holds across all datasets.

\begin{table}[t!]
  \centering
  \scriptsize
  \setlength{\tabcolsep}{4pt}
  \begin{tabular}{l r@{\hskip 10pt} l r}
    \toprule
    Matcher & T\,(s) & Matcher & T\,(s) \\
    \midrule
    XFeat-Star             & 0.3 & DeDoDe-LG         & 3.8 \\
    XFeat                  & 0.3 & RoMaV2            & 4.1 \\
    Tiny-RoMa              & 0.3 & MA-ELoFTR         & 4.7 \\
    MINIMA-RoMa-Tiny       & 0.4 & DUSt3R            & 4.9 \\
    XoFTR                  & 0.4 & RoMa              & 5.0 \\
    SIFT-NN                & 0.4 & MINIMA-XoFTR      & 5.3 \\
    SuperPoint-LG          & 0.4 & MINIMA-RoMa       & 5.5 \\
    ALIKED-LG              & 0.5 & GIM-DKM           & 5.5 \\
    DISK-LG                & 0.9 & LoFTR             & 5.6 \\
    GIM-LG                 & 0.9 & RoMa+LoFTR        & 6.7 \\
    SIFT-LG                & 1.4 & RoMa+Tiny-RoMa    & 6.8 \\
    MASt3R                 & 2.5 & OmniGlue          & 11.5 \\
    \bottomrule
  \end{tabular}
  \caption{\textbf{Per-pair runtime on SpaceNet9 (RTX~3090).} End-to-end
    wall-clock per pair (tile extraction, matcher forward pass, and RANSAC
    fit) under the Table~\ref{tab:final-master-ranking} protocol
    (1024\,px max side, $512{\times}512$ tiles with 256\,px overlap,
    $\sim$4 tiles per pair), averaged over all labeled SpaceNet9 pairs.
    Sorted by time within each column; matchers span more than an
    order of magnitude. Preprocessing and I/O are excluded as they are
    essentially constant across matchers.}
  \label{tab:runtime}
\end{table}

\custompar{Practical deployment guidance.}
Practitioners can achieve strong optical--SAR registration with off-the-shelf matchers.
Our ablations point to a concrete protocol baseline:
(i)~affine geometry over homography; (ii)~$512\times512$ tiles with
256\,px overlap; (iii)~per-matcher normalization selected on a small
labeled subset (Z-Score or percentile for the top matchers,
Table~\ref{tab:final-master-ranking}); (iv)~a RANSAC threshold of
3\,px for strong dense matchers, more permissive ($t{\geq}5$\,px) for
sparser matchers; and
(v)~MINIMA-RoMa or RoMa as the first-choice matcher when accuracy and
cross-dataset consistency dominate, or XoFTR when per-pair latency
matters (${\approx}0.4$\,s vs.\ $5.0$\,s per pair at equal SpaceNet9 accuracy).
Tile overlap and minimum inlier count are comparatively insensitive and
do not require further tuning.
This baseline achieves $<$8\,px mean error on SpaceNet9 and $47.0$\,px on SRIF
without any domain-specific training.

\custompar{Computational cost.}
Table~\ref{tab:runtime} reports per-pair end-to-end wall-clock on a
single NVIDIA RTX~3090 under the Table~\ref{tab:final-master-ranking}
protocol. Runtime ranges from $0.3$\,s (XFeat-Star) to $11.5$\,s (OmniGlue),
more than an order of magnitude.
Higher-accuracy RoMa variants and ensembles are slower. The cheapest detectors
are least accurate.
XoFTR is the exception. It sits in the top
accuracy tier on Table~\ref{tab:final-master-ranking} at ${\approx}0.4$\,s per
pair, roughly an order of magnitude faster than the RoMa family. It is the
strongest choice when per-pair latency matters.

\section{Limitations}
\label{sec:limitations}
\custompar{Scene and annotation coverage.}
SpaceNet9 train provides only three labeled scenes with residual
annotation and geolocation noise; scene-level statistical power is
limited despite sweeping 64~protocol configurations per matcher.
Results should be treated as diagnostic evidence at the pair and tile
level, not statistically independent scene samples.

\custompar{Matcher coverage and adaptation scope.}
We evaluate twenty-four families available through \texttt{vismatch} at
the time of experimentation; MapGlue~\citep{wu2026soma1m} and other
recently released matchers are not yet integrated.
We evaluate only pretrained matchers without adaptation. Lightweight
fine-tuning (e.g., adapter modules or modality-specific pretraining
as in XoFTR~\citep{tuzcuoglu2024xoftr}) may close the gap
for fragile matchers like MASt3R and DUSt3R.

\section{Conclusion}
We present a systematic zero-shot evaluation of twenty-four pretrained matcher
configurations on optical--SAR registration across SpaceNet9, SRIF, and
SARptical, without any weight adaptation.
On the three labeled SpaceNet9 training scenes, XoFTR and RoMa tie for the
lowest mean tie-point error in our sweep ($3.0$\,px), with MA-ELoFTR
($3.4$\,px) close behind; XoFTR reaches this accuracy at ${\approx}0.4$\,s per
pair, roughly an order of magnitude faster than the RoMa family.
Because normalization is selected on the same scenes, these results are
descriptive rather than held-out generalization estimates.
Across the evaluated ablations, geometry model and RANSAC settings change
performance by amounts comparable to---and sometimes exceeding---differences
between matchers, up to $33\times$ in mean error for a single matcher.
Performance does not transfer uniformly across datasets or metrics:
MINIMA-RoMa attains the best SARptical AUROC and the strongest zero-failure
cross-dataset consistency, MA-ELoFTR the best SARptical Recall@1, and SRIF
rankings depend strongly on failure handling.
Pretrained matchers thus provide useful zero-adaptation baselines for
optical--SAR registration, but reliable deployment requires explicit protocol
selection and cross-dataset validation.

\FloatBarrier
{\small
  \bibliographystyle{ieeenat_fullname}
  \bibliography{references}
}

\end{document}